\documentclass{article} 
\usepackage{collas2024_conference,times}
\usepackage{easyReview}

\usepackage{amsmath,amsfonts,bm}

\def\eqref#1{equation~\ref{#1}}

\def\1{\bm{1}}

\DeclareMathAlphabet{\mathsfit}{\encodingdefault}{\sfdefault}{m}{sl}
\SetMathAlphabet{\mathsfit}{bold}{\encodingdefault}{\sfdefault}{bx}{n}

\usepackage{hyperref}
\hypersetup{
    colorlinks=true,
    linkcolor=red,
    filecolor=magenta,
    urlcolor=blue,
    citecolor=purple,
    pdftitle={Overleaf Example},
    pdfpagemode=FullScreen,
    }

\usepackage[framemethod=TikZ]{mdframed}
\usepackage{enumitem}

\title{Reflecting on the State of Rehearsal-Free\\Continual Learning with Pretrained Models}

\collasfinalcopy

\author{Lukas Thede$^{1,2,*}$, Karsten Roth$^{1,2,*}$, Olivier Hénaff$^{3}$, Matthias Bethge$^{1}$, Zeynep Akata$^{2,4}$\\
$^1$T\"ubingen AI Center, University of T\"ubingen $^2$Helmholtz Munich, MCML $^3$Google DeepMind, $^4$TU Munich\\
$^*$equal contribution
}

\usepackage{microtype}
\usepackage{graphicx}
\usepackage{subfigure}
\usepackage{booktabs} 

\usepackage{amsmath}
\usepackage{amssymb}
\usepackage{mathtools}
\usepackage{amsthm}
\usepackage{multirow}
\usepackage{wrapfig}

\usepackage[capitalize,noabbrev]{cleveref}
\usepackage{xcolor}
\usepackage{colortbl}

\theoremstyle{plain}

\theoremstyle{definition}

\theoremstyle{remark}

\begin{document}


\maketitle

\begin{abstract}
With the advent and recent ubiquity of foundation models, continual learning (\textbf{CL}) has recently shifted from continual training from scratch to the continual adaptation of pretrained models, seeing particular success on rehearsal-free CL benchmarks (\textbf{RFCL}).
To achieve this, most proposed methods adapt and restructure parameter-efficient finetuning techniques (\textbf{PEFT}) to suit the continual nature of the problem.
Based most often on input-conditional \textit{query-mechanisms} or \textit{regularizations} on top of prompt- or adapter-based PEFT, these PEFT-style RFCL (\textbf{P-RFCL}) approaches report peak performances; often convincingly outperforming existing CL techniques.
However, on the other end, critical studies have recently highlighted competitive results by training on just the first task or via simple non-parametric baselines.
Consequently, questions arise about the relationship between methodological choices in P-RFCL and their reported high benchmark scores.
In this work, we tackle these questions to better understand the true drivers behind strong P-RFCL performances, their placement w.r.t. recent first-task adaptation studies, and their relation to preceding CL standards such as EWC or SI.
In particular, we show: \textbf{(1)} P-RFCL techniques relying on input-conditional query mechanisms work not because, \textit{but rather despite them} by collapsing towards standard PEFT shortcut solutions. 
\textbf{(2)} Indeed, we show how most often, P-RFCL techniques can be matched by a simple and lightweight PEFT baseline.
\textbf{(3)} Using this baseline, we identify the \textit{implicit bound on tunable parameters} when deriving RFCL approaches from PEFT methods as a potential denominator behind P-RFCL efficacy.
Finally, we \textbf{(4)} better disentangle continual versus first-task adaptation, and \textbf{(5)} motivate standard RFCL techniques s.a. EWC or SI in light of recent P-RFCL methods.
Together, we believe our insights to contribute to a more grounded treatment of CL with pretrained models.
\end{abstract}

\section{Introduction}\label{subsec:introduction}

While deep networks are widely adopted in research and application, the ability to continuously learn from novel contexts is still a challenging frontier yet to be surpassed \citep{kirkpatrick_2017_ewc,boschini2022xder,ostapenko2022foundation}. To tackle this, Continual Learning (\textbf{CL}) introduces extensive methodology; from weight \citep{kirkpatrick_2017_ewc} and gradient regularization \citep{chaudhry2019efficient} to rehearsal mechanisms \citep{boschini2022xder} which retain a memory of encountered data, and structural constraints \citep{mallya_2018_piggyback} - all to minimize catastrophic forgetting \citep{kirkpatrick_2017_ewc}, wherein learned model features are overwritten during continual training.


With ubiquitously available foundation models \citep{bommasani2021foundation}, focus in CL has, however, shifted away from continually learning models from scratch to continual adaptation of large-scale pretrained models (\textbf{PTMs}) to solve standard CL tasks, often entirely rehearsal-free (\textbf{RFCL}).
While works s.a \cite{ostapenko2022foundation,stojanovski2022momentumbased,zhang2023slca,roth2024fantastic} show success in continual fine-tuning and adaptation of entire pretrained models, most approaches derive from parameter-efficient finetuning (\textbf{PEFT}, ~\citet{jia2021visual,hu2021lora,kopiczko2024vera}). 
Such PEFT-style RFCL (\textbf{P-RFCL}) augments prompt- or adapter-based PEFT methods to account for the non-stationary data distribution; with a large body of work relying on input-conditional query-mechanisms or regularization on top of PEFT models~\citep{wang2022l2p,wang2022dualprompt,smith2023coda,gao2023lae}.

However, while reported benchmark performances increased, recent works \citep{janson2022simple,panos2023first,gao2023lae,zhou2023revisiting,mcdonnell2023ranpac} have raised first questions regarding the interpretation of these results and the overall formulation of the problem setting.
For example, \citet{gao2023lae,zhou2023revisiting} presented initial evidence on how query-based P-RFCL methods are often matched and even outperformed when combining adapter-based PEFT methods with general extensions such as ensembling or concatenation of multiple model outputs on top.
These approaches do not explicitly account for the continual nature of the problem - something deemed crucial in initial P-RFCL works~\citep{wang2022l2p,wang2022dualprompt}.
On top of that, works such as \citet{janson2022simple,panos2023first,zhou2023revisiting} have raised red flags w.r.t. the actual benchmark results for P-RFCL methods. 
More precisely, these works highlight how simple adaptation on \textit{only the first task} followed by subsequent gradient-free model alignment (e.g. through the use of a nearest mean classifier head) can already achieve competitive performance. 

Together, \textit{loose threads and open questions arise} regarding our understanding of P-RFCL and its practical utility to satisfy standard CL desiderata; facilitating learning along the datastream while mitigating catastrophic forgetting.
In this work, we contribute a sequence of experimental studies reciprocating and extending previous research to facilitate better, more reflective understanding of P-RFCL via contextualization and insights into these \textcolor{blue}{\textbf{questions}}:
\begin{mdframed}[roundcorner=5pt,backgroundcolor=blue!5,linewidth=1pt,innertopmargin=8pt,innerbottommargin=8pt]
\begin{enumerate}[leftmargin=10pt,label=\textcolor{blue}{\textbf{\Alph*.}}]
    \item \textit{How important are input-conditional query mechanisms in adapting standard parameter-efficient finetuning to rehearsal-free continual learning with pretrained models?}
    \item \textit{More generally, what drives strong performance of P-RFCL methods on common CL benchmarks?}
    \item \textit{How can P-RFCL be motivated in light of competitive first task adaptation methods?}
    \item \textit{What importance do standard rehearsal-free continual learning methods such as EWC or SI have in the age of P-RFCL and ubiquitous (large-scale) pretrained models?}
\end{enumerate}
\end{mdframed}
\vspace{-5pt}

These questions probe the methodological subset that popularised P-RFCL (input-conditional query-mechanisms~\citep{wang2022l2p,wang2022dualprompt}), as well as the general domain of P-RFCL and its placement in current and past RFCL literature.
To answer these questions, we provide experimental studies that lead us to the following conclusions:

\textbf{\textcolor{blue}{[A]} Query-based P-RFCL methods work by collapsing towards simple PEFT} (Section~\ref{subsec:query_collapse}).\\ 
We probe query-based P-RFCL methods in detail and reveal a prompt collapse shortcut \citep{gheiros2020shortcut} when trained naively, wherein query-based approaches encourage highly similar prompts to be learned and queried, converging towards simple prompt-tuning. Experiments using both a removed and an oracle query function further reveal that query-based P-RFCL works not despite,\textit{ but precisely because of a collapse} of the query function. 
Moreover, explicitly leveraging task-specific context and moving away from simple PEFT tends to \textit{incur a drop} in performance for P-RFCL.
This means that in practice, there is much less difference (and incentive to differentiate) between query-based P-RFCL systems and their underlying PEFT baselines; moving these different branches much closer together.

\textbf{\textcolor{blue}{[B]} Most P-RFCL methods are matched by simple PEFT, and variations on top are often superfluous} (Section~\ref{subsec:uniprompt}).\\ 
We show experimentally that under fair comparisons, there is no free lunch with respect to different query- and adapter-based P-RFCL approaches. 
Indeed, a simple prompt-tuning PEFT baseline (\textit{OnlyPrompt}) mimicking the collapsed limit-case of query-based P-RFCL can match or outperform existing P-RFCL approaches with much fewer parameters.
Alongside similar method performances, the efficacy of \textit{OnlyPrompt} strongly points towards parameter-efficient finetuning as the crucial, shared driver in performance, and that variations on top often provide only diminishing returns.

\textbf{\textcolor{blue}{[B]} Prompt-based PEFT operates in benchmark-beneficial regimes via tunable parameter count choices} (Sec.~\ref{subsec:uniprompt}).\\ 
As simple PEFT seems to be crucial for P-RFCL, we study \textit{why} it works so well for standard CL tasks using our \textit{OnlyPrompt} studycase. We find that the often default and intuitive choice of tunable parameter counts provides a natural trade-off between model forgetting and knowledge gain. This means that beyond any explicit methodological choice, it is the implicit choice of reducing the number of adaptable parameters that gives high CL benchmark scores. This stands in contrast to continual learning from scratch, where such parameter-count regularization is much less possible. Finally, we show that many prompt-based P-RFCL approaches fall within optimal parameter count bands found in our study.

\textbf{\textcolor{blue}{[C]} P-RFCL still has merits despite recent first task adaptation works} (Section~\ref{subsec:full_vs_single_adapt}).\\ 
We reciprocate and augment surprising results found in first-task adaptation works \citep{zhou2023revisiting,panos2023first}, but also show how, across extensive benchmark and task sequence comparisons, there is often still a significant benefit in conducting P-RFCL over the entire datastream. For robust comparisons, future works should explicitly account for benchmark-specific susceptibility towards calibration issues, which disproportionately affect parametric methods.

\textbf{\textcolor{blue}{[D]} Standard regularization-based CL can improve performance and robustness for P-RFCL} (Section~\ref{sec:pushing_limits}).\\ 
We leverage \textit{OnlyPrompt} to revisit the relevance of regularization-based CL methods. These are often negatively portrayed in P-RFCL works, which often report significant relative performance gains. While this appears to "fast-track progress (...) for CIL" \citep{marouf2023rethinking}, we show that these standard approaches still retain their core validity, with PEFT-style P-RFCL providing not an alternative but an orthogonal approach to CL. On top of these, methods like EWC or SI can operate well, improving both performance and robustness towards parameter count choices.

Altogether, this work aims to provide a clearer perspective on the status of P-RFCL by bringing together divergent branches within P-RFCL, providing a better understanding of underlying assumptions; highlighting and understanding simple PEFT as the primary driver in performance; and more clearly bridging P-RFCL with standard CL.
In doing so, we hope to provide a strong foundation for more grounded research into P-RFCL.

\section{Related Works}
\textbf{Continual Learning} (CL) studies the mitigation of catastrophic forgetting \citep{rusu_2016_pnn,kirkpatrick_2017_ewc,lopez2017gradient,chaudhry2019efficient} when training models under data- and task-streams with non-stationary, often continuous distribution shifts. It is often broadly classified into three main branches:
%
\textit{Regularization-based} approaches constrain learning on a weight level~\citep{kirkpatrick_2017_ewc,zenke2017continual,aljundi2018memory}). These preserve parameters deemed crucial for previous tasks while learning on subsequent ones. Effective in simpler settings, performances often degrade with complex datasets and -streams \cite{goodfellow2013empirical,kemker2018measuring,van2019three}.
%
\textit{Replay-based} methods maintain a memory of encountered samples to rehearse old context during task adaptation \citep{lopez2017gradient,chaudhry2019efficient,buzzega2020dark,prabhu_2020_gdumb,Cha_2021_co2l,boschini2022xder}, leveraging knowledge distillation \citep{hinton2015distilling} and self-supervised learning \citep{doersch2015unsupervised,gidaris2018unsupervised} to retain context from previous tasks. These methods generally perform favorably across benchmarks \citep{van2019three,delange2021review}, but can be slow or compromised by limited buffer sizes and unsuitable under data privacy and storage requirements \citep{shokri2015privacy}.
%
\textit{Architecture-based} methods modify architectures to accommodate continual learning; introducing task-specific components \citep{rusu_2016_pnn,yoon2017lifelong,golkar2019continual,wortsman2020supermasks} or task-dedicated sub-networks \citep{fernando2017pathnet,mallya_2018_piggyback}; but often require task identities at test time. Some~\citep{lee2019meta,rebuffi2017icarl,aljundi2019gradient} infer these but often introduce significant additional parameters, sometimes doubling model sizes \citep{fernando2017pathnet,mallya2018packnet}.

\textbf{Parameter-Efficient Finetuning (PEFT)} refines large pretrained models for downstream tasks without the overhead of full finetuning. 
While initial approaches have looked into specific finetuning protocols and additional parameter modules~\citep{houlsby2019parameter,pfeiffer_adapterfusion_2020,guo2021parameterefficient}, recent adapter-based approaches s.a. Low-Rank Adaptation~\citep{hu2021lora,zhang2023adaptive,lv2023adalomo,kopiczko2024vera} and Orthogonal Finetuning~\citep{qiu2023oft,liu2023parameterefficient,bini2024ether} have found success by learning weight modifiers which can be absorbed at test time. 
\citet{he2021towards} employs residual vectors to modify attention and feed-forward weights, while \citet{zaken2022bitfit,demin2023layernorm} only tune e.g. bias weights or layer norms.
Prompt- and Prefix-Tuning~\citep{lester2021power,li2021prefix} inserts learnable tokens into Transformer-style architectures to steer the model focus towards task-relevant features. 
Using Vision Transformers~\citep{dosovitskiy2020image,touvron2021training}, adaptations like Visual Prompt Tuning (VPT, ~\citet{jia2021visual}) or AdapterFormer~\citep{chen2021adapterformer} transfer these approaches to the visual domain.

\textbf{Rehearsal-free Continual Learning with Pretrained Models} explores pretrained models as knowledge bases to be utilized during adaptation to a datastream. 
Works such as \citet{stojanovski2022momentumbased} or \citet{zhang2023slca} show that interpolation or selectively reducing the learning rate in combination with posthoc alignment of the classification layers can mitigate catastrophic forgetting when continuously adapting pretrained backbones. 
At the same time, a large corpus of works has started exploring continual adaptation mechanisms by extending established PEFT approaches (like prompt tuning ~\citep{ke2020continual,wang2020efficient}), such as  \citet{wang2022l2p,wang2022dualprompt,smith2023coda,khan2023language,wang2023hierarchical} which have reported large performance gains by developing methods to correctly prompt PTMs.
More recently, approaches such as LAE~\citep{gao2023lae} or \citet{zhou2023revisiting} have shown comparable performance by directly utilizing PEFT objectives (while mitigating forgetting through interpolation or concatenation) without any query mechanisms.
Both \citet{smith2023coda}  and \citet{gao2023lae} have also raised first questions on the purpose of the (frozen) query mechanism behind, e.g., L2P or DP, though both works do not study these in detail. In this work, we provide an in-depth perspective on this mechanism and its relation to the direct PEFT methods in P-RFCL.
At the same time, works such as \citet{janson2022simple,panos2023first,zhou2023revisiting,mcdonnell2023ranpac} have shown that P-RFCL objectives on standard CL benchmarks can often be matched in performance by approaches adapting at most on the first task (with some non-parametric alignment such as Nearest Mean Classifiers for the remaining tasks).
Overall, in this work, we aim to connect all these diverging perspectives on P-RFCL with PTMs and provide stronger motivation and grounding for future work in this domain.

\section{Experimental Design and Notation}\label{sec:details}
We first provide a summary of experimental details shared across all experiments conducted in this work (\S\ref{subsec:experimental_details}), followed by an introduction and summary of relevant (query-based) P-RFCL approaches and respective notation (\S\ref{subsec:preliminaries}).

\subsection{Experimental Details}\label{subsec:experimental_details}

All experiments conducted in this paper utilize PyTorch~\citep{pytorch}. To ensure fairness and comparability of results, we transfer and implement all studied methods within the same repository, extending the repository provided in \citet{gao2023lae} with a common data loading and processing pipeline adapted from \textit{continuum}~\citep{douillardlesort2021continuum}. Method-specific hyperparameters follow the best settings provided in the respective original publications; however, orthogonal training parameters such as the number of iterations, learning rates, batch sizes, or the use of, e.g., weight decay, are adjusted to allow for direct comparability. Note that all methods that utilize a linear projection head leverage task masking as commonly done in P-RFCL, where task-unrelated logits are always masked \textit{during training}. We make use of the implementation provided in \citet{gao2023lae}.
Following existing P-RFCL publications (c.f. e.g. \citet{wang2022l2p,wang2022dualprompt,zhou2023revisiting,smith2023coda,khan2023language}, we \textit{start from} a \textit{timm}~\citep{timm} Vision Transformer (ViT) backbone with a patch size of 16 and weights pretrained on ImageNet-21k. All experiments are run on a high-performance compute cluster comprising NVIDIA 2080Tis. 

Image benchmarks covered in this paper include sequential variants of CIFAR100~\citep{cifar100}, ImageNet-R~\citep{imagenetr}, ImageNet-A~\citep{imageneta}, DomainNet~\citep{domainnet}, ObjectNet~\citep{objectnet}, OmniBench~\citep{omnibench}, CUB-200-2011~\citep{cub200-2011}, CARS196~\citep{cars196} and Caltech256~\citep{caltech256}. For consistency across benchmarks, we convert all datasets into streams of ten tasks with balanced class counts per task (except for DomainNet and Cars due to the number of classes provided). For additional implementation details, please refer to the Supplementary. Code is available at \url{https://github.com/ExplainableML/ReflectingOnRFCL/}.

\subsection{Preliminaries}\label{subsec:preliminaries}

\textbf{Class-Incremental Continual Learning.} We follow the class-incremental framing of CL (\textbf{CIL}) as utilized across the large majority of P-RFCL works~\citep{wang2022l2p,wang2022dualprompt,smith2023coda,gao2023lae,zhou2023revisiting}. In this scenario, a model is trained sequentially on $T$ tasks $\mathcal{D}:=\{\mathcal{D}_1,\mathcal{D}_1,...,\mathcal{T}_T\}$. Each task consists of a subset of input samples $x_i^t\in\mathcal{X}_t\in\mathcal{X}$, and their respective labels $y_i^t\in\mathcal{Y}_t\in\mathcal{Y}$, where class subsets are non-overlapping across tasks, i.e. $\bigcup_{t\in \{1, .., T\}}\mathcal{Y}_t = \emptyset$. $\left(\mathcal{X}, \mathcal{Y}\right)$ denote the entire set of samples and labels available in $\mathcal{D}$.
CIL aims to train a model $f_\theta:\mathcal{X} \rightarrow \mathcal{Y}$ with parameters $\theta$ to predict a label $y = f_\theta(x_\text{test})$ for any new input $x_\text{test}$ where $y(x_\text{test})\in\mathcal{Y}$. 
Additionally, CIL assumes access to the task ID \textit{only during training}.

\textbf{Query-based P-RFCL} leverages particular mechanisms to query a large-scale pretrained base model for data presented continually. Without loss of generality, we describe the basic mechanisms underlying these approaches using Learning-to-Prompt (L2P, \citet{wang2022l2p}) as a guiding example. L2P proposes a pool of prompts where each prompt consists of a key and value pair. This prompt pool can be viewed as a memory of learned adaptations retrieved using some query module. Practically, \citet{wang2022l2p} (but also other works s.a. \citet{wang2022dualprompt,khan2023language}) simply utilize the same, but frozen pretrained base model, which projects the input into the prompt-key space. Based on the projected input, the top-n closest prompts are retrieved and prepended to the sequence of embedded patches: 
\begin{equation}
    x_p = [P_1; ...;P_N;x_e], 1 \leq N \leq M,
\end{equation}
where $P_j\in\mathbb{R}^{L_p \times D}$ is a single prompt with token lenght $L_p$ from the prompt pool $P=\{P_1,P_2,...,P_M\}$ with size $M$. The extended input sequence is then passed through the frozen transformer backbone and the classification head. The only learnable parameters in this approach are the prompts (with both their keys and values) as well as the weights of the classification head. The objective function of L2P consists of a cross-entropy classification loss and a surrogate loss, minimizing the distance between the matched keys and the input query. 
Put together, the query mechanism effectively allows for a separation of adaptation modules across tasks and, ideally, visual concepts to mitigate catastrophic forgetting.

\textbf{Parameter-Efficient Finetuning (PEFT)} adapts pretrained models while minimizing computational overhead on both tuning resources and tunable parameter counts. Most recent approaches build either (1) on top of adapter-style approaches following LoRA~\citep{hu2021lora}, which learn low-parametrized weight modifiers $\Delta{}W$, $(W+\Delta{}W)^{\intercal}x + b$, by e.g. framing $\Delta{}W = AB$ as the product of two low-rank matrices $A, B$, or prompt approaches~\citep{li2021prefix,lester2021power} which learn input modifiers by concatenating learnable tokens to the input of a transformer block or the parameters of attention modules, for example
\begin{equation}
    \hat{x}^m = \left[P_1; ...; P_N; \hat{x}_{p_1}, ... \hat{x}_{p_L}\right].
\end{equation}
where $\hat{x}^m$ denotes the modified input to some transformer block with $\hat{x}_{p_i}\in\mathbb{R}^d$ the respective $L$ input patches, and $P_k\in\mathbb{R}^d$ the $N$ learnable prompt tokens.

\section{Revisiting PEFT-based Rehearsal-Free CL with Pretrained Models}

\begin{table}[t]
    \centering
    \caption{\textbf{Ablating query functions in query-based P-RFCL.} Our results show that high-performing query-based P-RFCL approaches encourage high similarity between query-retrieved learned prompts, and favor convergence towards simple prompt-based PEFT with minimal diversity between query-retrieved learned prompts.}
    \resizebox{0.9\linewidth}{!}{%
    \begin{tabular}{clccccc}
        \toprule
        \multirow{2}{*}{\textbf{Type}} & \multirow{2}{*}{\textbf{Method}} & \multirow{2}{*}{\textbf{Params (K)}} & \multicolumn{2}{c}{\textbf{CIFAR100}} & \multicolumn{2}{c}{\textbf{ImageNet-R}} \\
         & & & Prompt Sim. & Accuracy & Prompt Sim. & Accuracy \\
        \midrule
        \multirow{3}{*}{pool} & L2P \citep{wang2022l2p} & 46 & 96.93 & 85.02 ($\pm$0.82) & 92.74 &  68.43 ($\pm$1.49) \\
        & $\vartriangleright$ oracle query & \multirow{2}{*}{$\downarrow$} & 93.18 &  84.92 ($\pm$0.67) & 92.21 &  67.46 ($\pm$1.84) \\
        & $\vartriangleright$ w/o query & & 100 & 86.47 ($\pm$1.35) & 100 & 69.36 ($\pm$1.80)\\                
        \multirow{2}{*}{pool} & Dual \citep{wang2022dualprompt} & 253 & 93.26 & 84.21 ($\pm$1.34) & 95.72 &  68.06 ($\pm$0.93) \\
        & $\vartriangleright$ oracle query & \multirow{1}{*}{$\downarrow$} & 91.73 &  84.40 ($\pm$0.99) & 95.30 &  67.78 ($\pm$1.18) \\
    
        \cmidrule(lr){1-7}
        \multirow{2}{*}{learned} & CODA \citep{smith2023coda} & 150 & 99.75 & 86.17 ($\pm$0.42) & 80.53 & 70.77 ($\pm$0.71) \\
        & $\vartriangleright$ oracle query & $\downarrow$ & 91.47 & 79.79 ($\pm$1.41) & 71.80 & 66.28 ($\pm$1.84) \\        
        \cmidrule(lr){1-7}
        task & HiDe \citep{wang2023hierarchical} & 384 & 88.49 & 85.30 ($\pm$0.78) & 93.63 & 75.78 ($\pm$0.78) \\
        \bottomrule
    \end{tabular}
    }
    \label{tab:query_collapse}
\vspace{-5pt}
\end{table}
In this section, we highlight our conducted experimental studies alongside respective key findings and their interpretation and implications for the field of PEFT-based rehearsal-free continual learning with pretrained models (P-RFCL).

\subsection{On the Importance of Querying in Query-based P-RFCL}\label{subsec:query_collapse}
The use of query mechanisms for P-RFCL serves as the conceptual translation from simple PEFT to the continual learning domain \citep{wang2022l2p,wang2022dualprompt}: By making parameter usage conditional on an input-dependent query process, knowledge integration should become a selective process, in which catastrophic forgetting is mitigated (by avoiding selection of parameters for irrelevant tasks) and forward transfer enabled (by selecting parameters relevant to a specific task). This section explores the importance of query mechanisms and their impact on adaptation in more detail.

\paragraph{What prompts are learned and assigned in query-based systems?} By re-routing or re-weighting training signal based on input context (such as class-, task- or concept-level properties), the query functions should encourage different prompts to be learned and associated with varying input samples~\citep{wang2022l2p,wang2022dualprompt,smith2023coda,khan2023language}. 
Consequently, we expect the query mechanism to assign a diverse set of prompts across the test samples covering the entirety of seen tasks.
More precisely, we measure this using the average retrieved prompt similarity
\begin{equation}
    \textstyle P_{\text{sim}} = \left(\frac{1}{K} \sum_{i=1}^{K} \text{sim}_{\text{cos}}(\vec{p}_i, \vec{p}_{\text{proto}})\right) \times 100,
\end{equation}
where $K$ denotes the number of test samples for a given benchmark and $P_\text{sim}$ effectively measures the average similarity of a retrieved learned prompts to the prompt prototype $\vec{p}_{\text{proto}} = \frac{1}{K}\sum_{i=1}^K \vec{p}_i$ - indicating the retrieved learned prompt average - with $\vec{p}_i$ the $i^{th}$ prompt, and $\text{sim}_{\text{cos}}(\cdot, \cdot)$ the cosine similarity. Note that $P_\text{sim}\in\left[0,100\right]$. For completeness, we also include the recent SOTA HiDe Prompt \citep{wang2023hierarchical}\footnote{To allow for a fair comparison, we include HiDE-Prompt without additional classifier head alignment but provide an additional ablation in the Supplementary (\S Table \ref{tab:tap_comp}).} which employs task-specific prompts to induce prompt separation, amongst other extensions. While this approach does not include a query function, it still allows us to study a collapse toward similar prompts.

Results for the two most commonly used benchmarks for P-RFCL~\citep{wang2022dualprompt,smith2023coda, gao2023lae} - CIFAR100 and ImageNet-R - are presented in \cref{tab:query_collapse}, providing different similarity scores across benchmarks and methods. In all cases, similarity scores go significantly up when compared to the randomly initialized prompts before training ($<30$). 
Across both benchmarks and methods, $P_\text{sim}$-scores are very high, and for the majority of benchmarks and default methods exceed $90$ (e.g. $96.93$ for L2P and $99.75$ for CodaPrompt on CIFAR100, or $93.63$ for HiDE-Prompt on ImageNet-R). With $100$ indicating the same prompt being retrieved and used for every test sample, the results clearly indicate that by default, these query-based approaches closely converge to a collapsed query function, in which only the same prompt is retrieved - despite a large prompt budget in all cases.
These high prompt retrieval similarities are also reflected in high redundancy in the overall set of learned prompts: For L2P, the number of learned prompts can be noticeably reduced by, e.g., $30\%$ (pruned randomly), with performance only changing marginally ($85.02\%\rightarrow{}84.71\%$).

\paragraph{How important is a good query function?} We further conduct a simple experiment in which we extend the conceptual motivation behind the use of a query function:
On a high level, the success of the query mechanism rides on the ability to meaningfully separate samples across the entire datastream from the start, for which all query-based P-RFCL approaches leverage a separate pretrained model (usually the same backbone that is also adapted).
Therefore, we simply replace the default query function with the backbone finetuned on the entire datastream in an iid-fashion beforehand, thereby introducing an oracle query that has clear knowledge about the expected datastream ahead of time. 
As a result, any input-conditioning and separation of tunable features should be much stronger and consequently drive higher performance gains. 
We test this for representative query-based approaches using either prompt pools (L2P~\citep{wang2022l2p}, DualPrompt~\citep{wang2022dualprompt}) or end-to-end optimization over prompts (CodaPrompt~\citep{smith2023coda}). Results are again displayed in \cref{tab:query_collapse}.

As expected, the average retrieved prompt similarity $P_\text{sim}$ decreases in most cases, albeit sometimes only marginally - as even with a much better-suited, stronger initial context separation through the finetuned query function, all methods still appear to encourage a query function collapse.
Surprisingly, however, for a large majority of instances, a better query function and thus decreased prompt similarity actually drops performance in parts noticeably. For example, on ImageNet-R, we find an accuracy drop from $68.43\%$ to $67.46\%$ for L2P, $68.06\%\rightarrow{}67.78\%$ for DualPrompt and $70.77\%\rightarrow66.28\%$ for CodaPrompt. On CIFAR100, the drop can become even more significant, going from $86.17\%$ to $79.79\%$ for CodaPrompt. We attribute the drop in performance to an increase in forgetting as we observe the tendency of more specific adaptations to overfit to the samples from the latest task, overwriting adaptations to previous ones.

These aspects are further highlighted when getting rid of the entire query function in L2P to have no prompt separation of any form (i.e. an average retrieved learned prompt similarity of $P_\text{sim} = 100$), thereby effectively conducting simple prompt-based PEFT while achieving additional performance gains compared to its query-based variant (e.g. $85.02\%$ versus $86.47\%$ on CIFAR100).


\paragraph{\textcolor{blue}{Implications:}} These results provide the first experimental indication that the query function's practical role is likely misaligned with its conceptual motivation. The use of query mechanisms as a means to learn a meaningful separation and grouping of prompts for a continual adaptation problem has very little, in part detrimental practical impact on the final performance. Instead, query-based approaches appear to work particularly because, in practice, the utilized query mechanism collapses to not encourage meaningful prompt separation but rather gives rise to shortcuts in which very similar or even the same prompts are used throughout. This holds also when separating prompts based on task identities. And indeed, the highest performance is often achieved when minimizing the impact of the query mechanism either through end-to-end optimization or simply getting rid of the entire query mechanism and just directly conducting simple prompt-based PEFT.

\subsection{Parameter-Efficient Finetung for Rehearsal-Free Continual Learning}\label{subsec:uniprompt}
\begin{table*}[t]
    \centering
    \caption{\textbf{Method comparison} of different query-based approaches as well as adapter-based ensembling (LAE) across nine different continual learning streams build over the respectively mentioned datasets. Results reveal similar performances, which is often matched or even outperformed by conducting simple prompt-tuning based PEFT (\textit{OnlyPrompt}).}
    \resizebox{\textwidth}{!}{%
    \begin{tabular}{lcccccccccc|c}
        \toprule
        \textbf{Adaptation} & \textbf{Params (K)} & \textbf{CIFAR100} & \textbf{ImageNet-R} & \textbf{ImageNet-A} & \textbf{Domainnet} &  \textbf{ObjectNet} & \textbf{Omnibench} & \textbf{CUB} & \textbf{Cars} & \textbf{Caltech256} & \textbf{Avg.}\\
        \midrule
        L2P & 46.1 &  85.0 ($\pm$0.8) & 68.4 ($\pm$1.5) & 43.4 ($\pm$0.5) & 65.0 ($\pm$0.2) & 56.0 ($\pm$0.1) & 65.3 ($\pm$0.4) & 72.6 ($\pm$1.7) & 43.0 ($\pm$5.6) & 91.8 ($\pm$0.3) & 65.8 ($\pm$6.1)  \\
        Dual & 253 & 84.2 ($\pm$1.3) & 68.1 ($\pm$0.9) & 49.6 ($\pm$0.6) & 60.8 ($\pm$0.4)  & 53.3 ($\pm$0.8) & 59.9 ($\pm$1.8) & 73.2 ($\pm$1.5) & 40.8 ($\pm$4.2) & 91.3 ($\pm$0.3) & 64.6 ($\pm$5.2) \\ 
        CODA & 150 & 86.2 ($\pm$0.4) & 70.8 ($\pm$0.6) & 46.3 ($\pm$0.5) & 65.2 ($\pm$0.3) & 57.6 ($\pm$0.3) & 63.1 ($\pm$2.4) & 73.2 ($\pm$4.6) & 41.5 ($\pm$4.6) & 92.7 ($\pm$0.3) & 66.3 ($\pm$7.0)\\
        HiDe & 384 & 85.3 ($\pm$0.8) & \textbf{75.8} ($\pm$0.8) & 44.7 ($\pm$1.7) & \textbf{66.0} ($\pm$0.2) & 59.9 ($\pm$0.8) & 67.9 ($\pm$1.0) & \textbf{82.7} ($\pm$0.6) & 47.4 ($\pm$11.4) & 94.3 ($\pm$0.1) & 69.3 ($\pm$11.7) \\
        \midrule
        LAE (adapter) & 84.5 & 85.8 ($\pm$0.6) & 72.6 ($\pm$0.7) & 51.6 ($\pm$0.6) & 63.4 ($\pm$0.3) & 54.3 ($\pm$0.9) & 60.7 ($\pm$1.7) & 74.8 ($\pm$0.6) & \textbf{50.5} ($\pm$5.7) & 90.9 ($\pm$0.8) & 67.1 ($\pm$6.2)\\
        \midrule
        \textit{OnlyPrompt} & 0.77 & \textbf{86.9} ($\pm$0.4) & 73.0 ($\pm$1.7) & \textbf{52.6} ($\pm$0.2) & 65.1 ($\pm$0.3) & \textbf{61.3} ($\pm$0.4) & \textbf{69.9} ($\pm$0.6) & 78.7 ($\pm$2.8) & 42.3 ($\pm$5.6) & \textbf{94.9} ($\pm$0.3) & \textbf{69.4} ($\pm$6.8) \\
        \bottomrule
    \end{tabular}}
    \label{tab:method_comp}
\vspace{-8pt}
\end{table*}

\paragraph{The \textit{OnlyPrompt} baseline.} As we find query-based approaches to practically (and surprisingly, favorably!) collapse towards simple, prompt-based PEFT, we want to extend this insight and understand the practical limits of simple PEFT for P-RFCL.
To do this, we introduce the \textit{OnlyPrompt} baseline - a clean and simple reference approach, which simply reflects the collapsed limit-case of L2P, and consequently follows simple prompt-based PEFT (c.f. Sec.~\ref{subsec:preliminaries}) of a pretrained and frozen backbone. More precisely, \textit{OnlyPrompt} prepends $N$ $D$-dimensional trainable tokens $\{P_i\}_{i\in[1,...,N]}$ to the input, and incorporates the standard task-masking of output logits during training. The only hyperparameter that is introduced on top is the relative change in learning rate $\lambda$ of the input tokens with respect to the linear classification head.

Of course, applying PEFT methods to Continual Learning is not novel and has been explored before in, e.g., \citet{zhou2023revisiting} for first-task adaptation alongside model concatenation, or \citet{gao2023lae} as a naive baseline to compare against. In \citet{gao2023lae}, it is subsequently outperformed by their ensemble variant LAE comprising both weight-averaged offline and an adaptive online model. 
Generally, much more expensive and complicated systems are established on top of these baselines for additional performance gains, increasing both the number of hyperparameters and inference cost (both \citet{gao2023lae} and \citet{zhou2023revisiting} require multiple forward passes).

The primary purposes of \textit{OnlyPrompt} are to reciprocate these baseline insights, study the L2P collapse limit case, but also specifically highlight how well \textit{simple} PEFT performs, beating both the more convoluted query-based approaches and the ensembling setup utilized in LAE.
Across many established and additional Continual Learning benchmarks, our experiments in \cref{tab:method_comp} first show how similar most approaches perform when compared fairly.
More importantly, however, it reveals how other established P-RFCL approaches are easily matched or outperformed. For example, \textit{OnlyPrompt} - which uses only a fraction of the parameters - improves on average over all nine benchmarks by more than $2pp$ over LAE and by more than $3pp$ over any query-based approach. In addition to that, it matches the much more complex SOTA HiDE-prompt.
However, it is not the absolute performance that should be treated as the main takeaway\footnote{With the choice of inductive biases in network design, more complex architectural extensions can help explicitly tackle shortcomings in pure PEFT on individual benchmarks such as ImageNet-R or OmniBench.}, but rather that across such small- to mid-scale, standardized benchmarks, most existing approaches are heavily over-parameterized, introducing a large number of potentially confounding elements that overshadow a potentially key driver in performance: simple parameter-efficient fine-tuning. 

\begin{figure*}
    \centering
    \includegraphics[width=\linewidth]{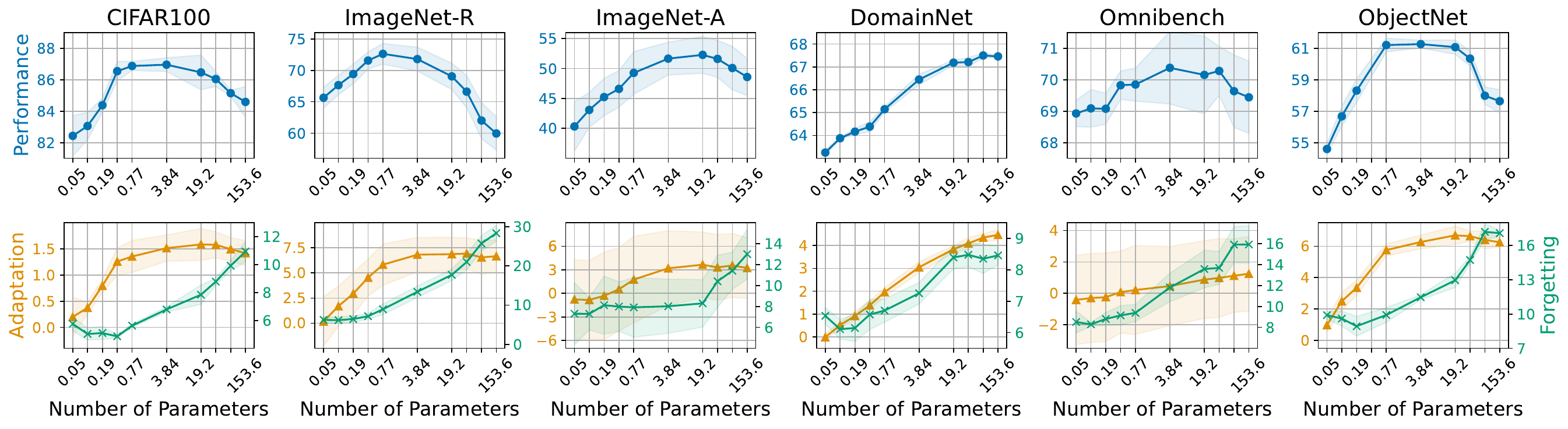}
    \vspace{-20pt}
    \caption{\textbf{Evaluation of performance versus number of tunable parameters} for PEFT approaches deployed to rehearsal-free continual learning. We use \textit{OnlyPrompt} as our easily controllable reference approach. Alongside the fact that most RFCL approaches can be boiled down to PEFT, our results indicate the number of tunable parameters as an implicit regularizer that allows for an effective tradeoff between adaptation and forgetting.}
    \label{fig:adaptation_forgetting}
\vspace{-10pt}
\end{figure*}

\paragraph{Why does PEFT work so well for P-RFCL?} Given the strong results of \textit{OnlyPrompt} and the fact that query-based methods operate well because they collapse towards simple PEFT, the question naturally arises as to why PEFT lends itself so well to these standard CL benchmarks.

Since the different P-RFCL methods (c.f. \cref{tab:method_comp}) perform similarly, and simple prompt-tuning via \textit{OnlyPrompt} achieves such good results, there is little evidence that it is the exact style of parameter-efficient finetuning that makes or breaks the success of a particular approach to the CL scenario. The studied query-based approaches (albeit with query collapse, c.f. \cref{subsec:query_collapse}) vary heavily in how and where in the pretrained model they utilize their tunable parameter budget while performing similarly to the adapter-ensembling done in LAE.

Consequently, a key difference to full model finetuning lies within the overall number of parameters actually optimized for.
To investigate this aspect in detail, we conduct a parameter count versus performance study across \textit{all} benchmarks deployed in \cref{tab:method_comp} across more than four magnitudes. 
To avoid any confounders interfering with this study, we utilize \textit{OnlyPrompt}. This approach provides a simple and easily controllable interface to adjust the parameter count across arbitrary magnitudes. For higher parameter counts, we prepend additional prompts. In cases where the number of parameters is below that of a single prompt vector ($768$ for ViT-B/16), we achieve this by simply padding a lower-dimensional prompt vector.
Doing so gives rise to the performance plot in \cref{fig:adaptation_forgetting}. In addition to that, we also measure the amount of adaptation and forgetting\footnote{We quantify adaptation as the improvement in local task accuracy (i.e. without considering logits from other tasks) when adapting the backbone to said task compared to training only a linear probe. Forgetting is measured as the change in accuracy for a task, comparing its performance immediately after training with its performance after all subsequent tasks have been trained.} as a function of the same parameter counts in the row below.

As can be seen, across nearly every benchmark, there is an optimal parameter band in which a favorable trade-off between adaptation and forgetting can be struck.
This parameter band is very robust and can cover in parts multiple magnitudes for most datasets. This explains why so many variations of approaches can be proposed while offering comparable performance when fairly compared. For example, on CIFAR100 and ImageNet-R, this optimal parameter band falls within a [$190$, $19000$] interval - which incidentally also covers the chosen numbers of tunable \textit{\textbf{prompt}} parameters used for L2P, DualPrompt, and (variants of) CodaPrompt.

These results strongly indicate that it is not the choice of method but rather the implicit choice of limited tuning parameters (relative to the overall backbone model) that naturally regulates adaptation within an adaptation-forgetting optimal regime for rehearsal-free continual learning. This stands in contrast to normal continual learning from scratch, where given a fixed architecture choice, such a parameter-count-based regularization is not as straightforwardly done.

\paragraph{What drives PEFT optimality?} When looking at \cref{fig:adaptation_forgetting} more closely, we can see that inflection points for both parameter counts that are too low and too high, respectively, are in parts noticeably shifted between different benchmarks. As can be seen in \cref{fig:adaptation_forgetting}, the former is caused by an inherent lack of learned adaptation shared amongst all datasets. The latter, however, is driven by a different adaptation behavior while forgetting increases in a similar and fairly monotonous fashion with parameter counts.

\begin{wraptable}{r}{0.5\linewidth}
    \vspace{-15pt}
    \centering
    \caption{\textbf{FID between stratified ImageNet subsets and resp. datasets alongside inflection point estimates} based on \cref{fig:adaptation_forgetting}. We find little relation between a basic distribution shift and the optimal parameter adaptation count.}    
    \resizebox{1\linewidth}{!}{    
    \begin{tabular}{lcccc}
    \toprule
        \textbf{Dataset}       & IN-R & IN-A & D-Net & O-Net\\
        \textbf{FID}(ImageNet) & $55.7$ & $51.37$ & $81.22$ & $82.33$\\
        \textbf{Param. Inflection (K)} & $\approx 38.4$ & $>153.6$ & $>>153.6$ & $\approx 76.8$\\
    \bottomrule
    \end{tabular}}
    \label{tab:fids}
    \vspace{-5pt}
\end{wraptable}

Consequently, while within the same [$190$, $19000$] range of tunable parameters noted above, performance is somewhat consistently good (c.f. \cref{tab:method_comp}) across benchmarks, both optima and especially inflection points for maximal parameter counts are shifted across magnitudes.
This indicates that different benchmarks have different minimal and maximal context requirements that need to be (and can be) successfully adapted to.
However, as we find in \cref{tab:fids}, the optimal parameter band position and width is difficult to estimate \textit{a priori} - at the very least, simple data distributional distances (measured via FID~\citep{fid}) between pretraining and continual adaptation data has no predictive ability for the favorable number of tuning parameters needed. For example, both ImageNet-A and ImageNet-R, as well as DomainNet and ObjectNet, exhibit similar FID scores but vastly different parameter curve inflection points.
As such, high-level distributional distances to the pretraining data seem insufficient to explain why and when PEFT works well for continual adaptation.

\paragraph{\textcolor{blue}{Implications:}} Simple parameter-efficient finetuning for CL works well primarily because the implicit choice of tuning parameters naturally regulates adaptation within an adaptation-forgetting optimal regime for rehearsal-free continual learning. How these parameters are integrated has much less impact, as the comparable performance of approaches across benchmarks shows. 
However, the optimal parameter regime is difficult to estimate \textit{a priori}, with simple data distributional distances being insufficiently descriptive.

\subsection{On the Importance of Full Adaptation and the Choice of Benchmarks for P-RFCL}\label{subsec:full_vs_single_adapt}
\begin{table*}[t]
    \centering
    \caption{\textbf{First task adaptation versus full adaptation.} We showcase that a simple linear probe outperforms the recently proposed first task adaptation methods, with additional PEFT-style adaptation offering additional gains. Put together, our results show that there is still notable merit to conduct and research RFCL across the entire datastream. (***) denotes strongly divergent results across seeds for CUB and Cars.}
    \resizebox{\textwidth}{!}{%
    \begin{tabular}{lcccccccccc}
        \toprule
        \textbf{Adaptation} & \textbf{Params (K)} & \textbf{CIFAR100} & \textbf{ImageNet-R} & \textbf{ImageNet-A} & \textbf{Domainnet} &  \textbf{ObjectNet} & \textbf{Omnibench} & \textbf{CUB} & \textbf{Cars} & \textbf{Caltech256}\\
        \midrule
        Linear Probe & 0 & 83.1 ($\pm$0.7) & 66.0 ($\pm$0.9) & 49.8 ($\pm$0.6) & 63.7 ($\pm$0.2) & 55.1 ($\pm$0.1) & 70.3 ($\pm$0.4)  & 74.6 ($\pm$0.5) & 36.5 ($\pm$6.5) & 94.7 ($\pm$0.2)\\
        NMC & 0 & 80.7 ($\pm$0.1) & 53.3 ($\pm$0.1) & 45.3 ($\pm$1.1) & 45.2 ($\pm$0.0) & 51.2 ($\pm$0.1) & 73.3 ($\pm$0.2) & \textbf{85.1} ($\pm$0.2) & 35.6 ($\pm$0.2) & 93.3 ($\pm$0.1)\\
        \midrule        
        ADAM (VPT-s) & 3.84 & 80.6 ($\pm$1.2) & 54.3 ($\pm$1.0) & 27.8 ($\pm$4.7) & 46.5 ($\pm$1.3) & 47.7 ($\pm$3.1) & 73.3 ($\pm$0.2) & 84.9 ($\pm$0.4) & 29.7 ($\pm$6.0) & 93.6 ($\pm$0.1)\\
        RanPAC (VPT-s) & 3.84 & \textbf{87.0} ($\pm$0.8) & 69.8 ($\pm$1.2) & 47.43 ($\pm$3.2) & 63.4 ($\pm$0.4) & 57.9 ($\pm$0.5) & \textbf{77.2} ($\pm$1.3) & *** & *** & \textbf{95.1} ($\pm$0.1) \\
        \midrule
        \textit{OnlyPrompt} & 3.84 & \textbf{87.0} ($\pm$0.5) & \textbf{72.1} ($\pm$1.8) & \textbf{55.0} ($\pm$0.6) & \textbf{66.4} ($\pm$0.3) & \textbf{61.4} ($\pm$0.1) & 70.1 ($\pm$1.1) & 81.4 ($\pm$1.8) & \textbf{46.6} ($\pm$7.0) & 94.4 ($\pm$0.1) \\
        \bottomrule
    \end{tabular}}
    \label{tab:first_task}
\vspace{-8pt}
\end{table*}

In the previous sections, we have shown how query-based P-RFCL approaches in practice collapse towards simple parameter-efficient finetuning; working by implicitly operating within a benchmark-optimal parameter band. Particularly, controlled usage of simple prompt-based PEFT via \textit{OnlyPrompt} achieves convincing performance across benchmarks.
In this section, we aim to align these insights with recent publications claiming sufficiency of simple first task adaptation in P-RFCL (alongside some non-parametric alignment for the remaining tasks), such as \citet{zhou2023revisiting,panos2023first};\citet{mcdonnell2023ranpac}. In particular, we wish to understand if the performance insights and gains hold merit when accounting for the strong first-task-adaptation baselines addressed in these works.

\textbf{First task adaptation versus full adaptation - a comprehensive evaluation.} In particular, \citet{zhou2023revisiting} show how a simple nearest mean classifier (NMC) using the same pretrained backbone and without gradient updates can achieve good results; following which ADAM is introduced, which combines simple parameter-efficient adaptation on the first task and concatenation with the original pretrained model outputs for additional performance gains on top. Similarly, RanPAC \cite{mcdonnell2023ranpac} uses first task adaptation but additionally employs random projections to project the features into a higher dimension before utilizing them for NMC classification.
Our experiments follow the code provided in \citet{zhou2023revisiting} and \citet{mcdonnell2023ranpac} but are integrated within our benchmark setting for fair comparability, with results for all benchmarks provided in \cref{tab:first_task}. 

In our experimental analysis, we find that while NMCs perform admirably, even a simple linear classifier can consistently and notably outperform it (e.g. $66.0\%$ versus $53.3\%$ on ImageNet-R). More importantly, however, we find that such a simple linear probe adapted across the entirety of the datastream can reliably outperform various carefully designed ADAM variants described above: On ImageNet-R, a linear probe improves over the best ADAM variant by more than $10pp$, and on the challenging DomainNet by over $12pp$. These differences become even more exacerbated on some benchmarks when comparing against \textit{OnlyPrompt}, with gains up to $17pp$. On top of that, we find that when averaged across multiple different task sequences, first-task adaptation can vary heavily in performance. 
Overall, when evaluated across a large array of different benchmarks, first-task adaptation schemes fall short. This is particularly evident when accumulating results across different task sequences. Even more involved approaches like model concatenation underperform compared to a simple PEFT baseline adapted to the Continual Learning task at hand.

\textbf{A more detailed benchmark evaluation.} While the majority of results favor full adaptation, there is no free lunch. For some benchmarks, particularly OmniBench and CUB, we find the performance of \textit{OnlyPrompt} (but also all other P-RFCL approaches listed in \cref{tab:method_comp}), to both fall below ADAM and RanPAC, but more interestingly also below that of a simple NMC classifier. This gives strong validation towards the claims made in \citet{zhou2023revisiting,panos2023first}, questioning the usefulness of existing methods.
However, it also warrants a closer look, which shows that this phenomenon is closely tied to the performance delta between the linear probe and the NMC. Margins between the linear probe and the NMC go up to over $10pp$ in parts in these cases. As \textit{OnlyPrompt} leverages a linear probe for alignment to the classification tasks, this performance bias is further propagated.
A more detailed examination reveals that the linear classifier achieves superior average accuracies within individual tasks (local accuracies) on these benchmarks, exceeding the NMC head in parts notably. 
This finding is particularly significant because, in other benchmarks like CIFAR100, the linear classifier showed comparable advantages in both local and overall (global) accuracies. 

This phenomenon can be attributed to calibration issues and consequent biases in the linear probe towards more recent class, a well-studied problem~\citep{hou2019bias,zhao2020bias,mittal2021bias}. A consequence thereof is also a breakdown in effectiveness when different tasks introduce closely related, finegrained concepts. We find this to be the case for both OmniBench and CUB, where the classifier for one task also produces high confidence estimates for closely related variants in other tasks (e.g. \textit{swan} versus \textit{duck}). Regarding such calibration issues, first-task adaptation has the inherent benefit of stable feature representations throughout the task stream, favoring classifier alignment.
Consequently, it is important for future P-RFCL research to be informed about these benchmark differences when reporting results and potentially compare against confidence-calibrated variants when required.

\textbf{\textcolor{blue}{Implications:}} Importantly for P-RFCL with pretrained models, when investigated across a sufficiently large set of benchmarks and task orderings, adaptation across the entire datastream consistently and in parts heavily outperforms first task adaptation baselines. However, sufficient emphasis has to be placed on the exact benchmarks evaluated on and their receptiveness towards calibration issues in the last linear layer, which can skew results and interpretations.

\subsection{Standard Continual Learning Augments P-RFCL}\label{sec:pushing_limits}
\begin{wraptable}{r}{0.5\linewidth}
\vspace{-20pt}
    \centering
    \caption{\textbf{Standard CL regularizations for PEFT.} Applying established weight regularization techniques can often provide reliable performance gains.}
    \resizebox{0.9\linewidth}{!}{%
    \begin{tabular}{lccc}
        \toprule
        \textbf{Adaptation} & \textbf{CIFAR100} & \textbf{ImageNet-R} &  \textbf{ObjectNet} \\
        \midrule
        \textit{OnlyPrompt} & 86.96 ($\pm$0.54) & 73.03 ($\pm$1.66) & 61.39 ($\pm$0.07) \\
        \quad + EWC & \textbf{87.43} ($\pm$0.82) & \textbf{75.28} ($\pm$0.48) & \textbf{61.64} ($\pm$0.31) \\
        \quad + SI & 87.38 ($\pm$0.78) & 74.52 ($\pm$1.29) & 61.51 ($\pm$0.61) \\
        \bottomrule
    \end{tabular}}
    \label{tab:cl_methods}
\vspace{-5pt}
\end{wraptable}

Our experiments have concluded towards the primary driver behind P-RFCL approaches being the implicit choice of adaptation parameters when either explicitly choosing (or collapsing towards) simple PEFT, as opposed to any explicit method choices.
However, if P-RFCL on commonly tested benchmarks effectively boils down to the application of PEFT, where do these insights position within the larger continual learning literature?

\begin{figure*}
    \centering
    \includegraphics[width=\linewidth]{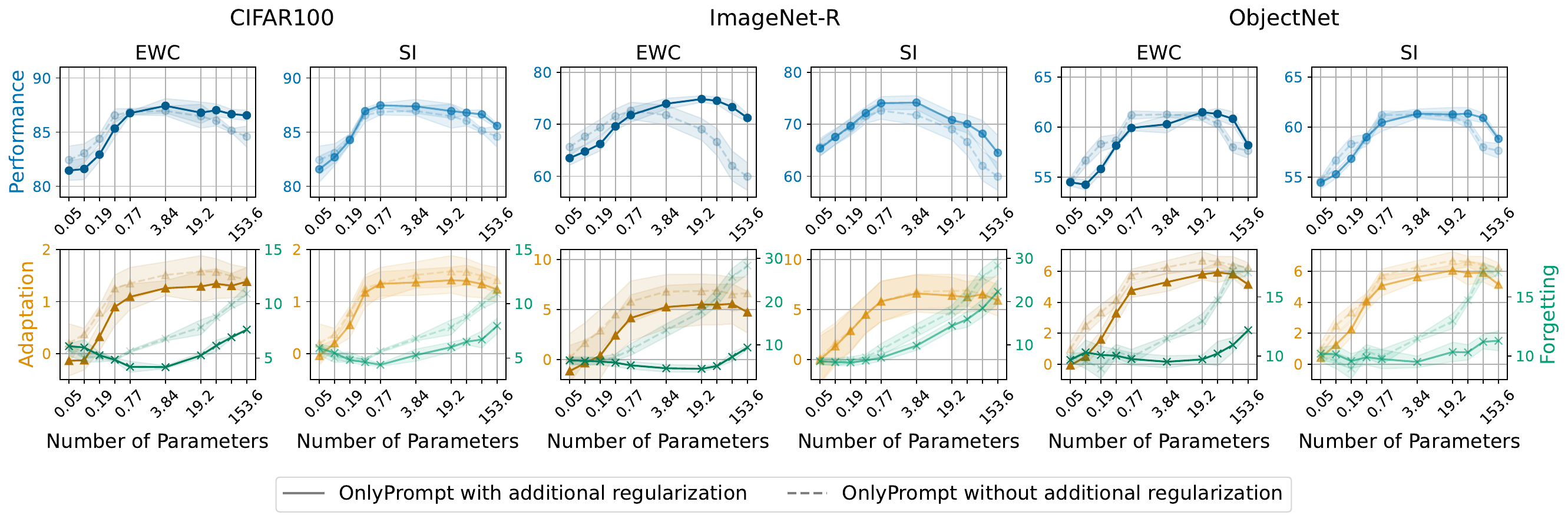}
    \vspace{-20pt}
    \caption{\textbf{Impact of standard weight regularization techniques on PEFT for CL.} When utilizing methods such as EWC or SI, we find the expected reduction of forgetting, as well as adaptation capacities to be diminished simultaneously. Put together, however, these methods both shift and elongate the optimal parameter band toward higher tunable parameter counts, making the result combination more robust toward parameter count choices.}
    \label{fig:fixprompt_regularization}
\vspace{-8pt}
\end{figure*}

Several papers advocate for P-RFCL as an alternative to standard approaches developed for continual learning of models from scratch, comparing P-RFCL variants against standard regularization objectives (e.g. \citet{wang2022l2p,wang2022dualprompt}) such as EWC~\citep{kirkpatrick_2017_ewc} or SI~\citep{zenke2017continual}. However, these approaches simply suggest weight constraints on the learnable parameters, and a complete claim should also study the orthogonality of these methods with respect to P-RFCL methods. As P-RFCL methods implicitly collapse towards simple PEFT or explicitly build on top of it, we utilize \textit{OnlyPrompt} as our representative base method, on top of which we apply either EWC or SI. Given the smaller number of parameters needed for adaptation, the use of either method is cheap (which is otherwise usually very resource-intensive; e.g. EWC requires a separate approximation to the Fisher Information Matrix to be stored).

Results for a large range of adaptation parameters are displayed in \cref{fig:fixprompt_regularization}, alongside best peak performances for \textit{OnlyPrompt} alone, with EWC and with SI in \cref{tab:cl_methods}. As can be seen on CIFAR100 and ImageNet-R, the application of regularization methods either retains or improves peak performance to a reasonable degree ($87.4$ with EWC versus $87.0$ without on CIFAR100 and $75.3$ versus $73.0\%$ on ImageNet-R). The exact impact of these methods is also clearly visible in \cref{fig:fixprompt_regularization}, which highlights a strong suppression of forgetting and the cost of overall adaptation strength.
Generally, however, the application of these methods shifts the high-performance parameter band towards favoring variations with higher parameter counts, as expected. Interestingly though, we find that this comes with either retention or extension (e.g., on CIFAR100 and ImageNet-R) of the optimal parameter bandwidth (c.f. \cref{fig:fixprompt_regularization}) when applied on top of \textit{OnlyPrompt}. As a consequence, we find that the usage of standard CL weight regularization techniques can often raise the performance of PEFT approaches for Continual Learning (and thereby P-RFCL methods) while increasing robustness towards initial parameter count choices.

\textbf{\textcolor{blue}{Implications:}} P-RFCL, particularly through the lens of PEFT, should not be treated as an alternative to standard weight regularization techniques but rather be seen as orthogonal. Applications of these methods on top of PEFT for CL give reliable gains in performance and robustness.

\section{Conclusion}
This paper studies important open questions in PEFT-based rehearsal-free continual learning with pretrained models (P-RFCL), relating to a lack of understanding regarding the inherent performance drivers behind P-RFCL methods, the relationships between existing methods, and the overall merits of conducting P-RFCL on standard CL tasks.
We first reveal that in practice, \textit{query-based P-RFCL methods collapse toward a simple PEFT} shortcut; circumventing the purpose of query mechanisms.
Surprisingly, it is precisely \textit{this collapse that facilitates high benchmark performance}, and a stricter focus towards actual query functionality brings down efficacy.
Our study also underscores the \textit{primary role of simple PEFT in driving performance} for P-RFCL on standard benchmarks: Many different P-RFCL approaches compare similarly, and a minimal PEFT-only reference method \textit{OnlyPrompt} can provide competitive, in parts superior performance compared to more complex counterparts. 
Using \textit{OnlyPrompt}, we show that the implicitly selected \textit{number of tunable parameters naturally imposes a favorable trade-off} between adaptation to new tasks and the forgetting of old knowledge. 
Our exploration also highlights \textit{retained merits of adaptation across the entire datastream}, augmenting recent first-task adaptation strategies. Across diverse benchmarks, we find clear benefits in adapting models to every task in a data stream, even in rehearsal-free continual learning \textit{with pretrained models}.
Lastly, we find conventional regularization techniques, such as Elastic Weight Consolidation (EWC) and Synaptic Intelligence (SI), to \textit{integrate well with P-RFCL methods}, bolstering both performance and robustness towards parameter count choices. 
This insight reasserts the enduring value of established continual learning strategies, even in an era dominated by CL with pretrained foundational models. 
We hope our findings can lay the groundwork for more principled future research in rehearsal-free continual learning with pretrained models, providing a clearer balance and re-use of PEFT and respective augmentation strategies. Finally, our results advocate for developing a broader benchmark spectrum to represent the diversity of practical use cases, which is essential for accurately assessing the efficacy of emergent P-RFCL methods.

\section*{Acknowledgements}
Lukas Thede and Karsten Roth thank the International Max Planck Research School for Intelligent Systems (IMPRS-IS) for
support. Karsten Roth also thanks the European Laboratory for Learning and Intelligent Systems (ELLIS) PhD
program for support. This work was supported by DFG project number 276693517, by BMBF FKZ: 01IS18039A,
by the ERC (853489 - DEXIM), by EXC number 2064/1 – project number 390727645. The project was made possible by funding from the Carl Zeiss Foundation.

\bibliography{main}
\bibliographystyle{collas2024_conference}

\newpage
\appendix

\Large{Supplementary Material}
\normalsize

This supplementary material provides additional information for our main paper, including a detailed look at Vision Transformers (ViT), benchmarks used for our experiments, and implementation details of the assessed methods.

\section{Vision Transformer Background Information}
In this section, we will provide a brief overview of the Vision Transformer (ViT) \citep{dosovitskiy2020image} architecture, which we use as the common backbone in our experiments. This architecture has gained a lot of attention in the field of computer vision in recent years and has been utilized as the pretrained backbone in various works studying the adaptation of pretrained models in the context of continual learning \citep{wang2022l2p, wang2022dualprompt, gao2023lae, zhou2023revisiting}.

\textbf{ViT Model Mechanics}: The ViT architecture dissects an input image into a sequence of non-overlapping patches, of fixed size (e.g. 16x16 pixels). These patches are linearly embedded into a high-dimensional space, akin to token embeddings in text applications, and a positional embedding is added to each to preserve their relative or absolute position in the input image. The mathematical representation of the embedding process is as follows:

\[
\text{PatchEmbeddings} = [\text{vec}(p_1)E; \text{vec}(p_2)E; ... ; \text{vec}(p_N)E] + E_{\text{pos}},
\]

where \( \text{vec}(p_i) \) is the flattened vector of the \( i \)-th image patch, \( E \) is the embedding matrix, \( E_{\text{pos}} \) is the positional embedding matrix, and \( N \) is the total number of patches.

Following embedding, the model applies the self-attention mechanism within the transformer blocks, allowing it to focus dynamically on different parts of the image. The attention operation can be succinctly expressed as:

\[
\text{Attention}(Q, K, V) = \text{softmax}\left(\frac{QK^T}{\sqrt{d_k}}\right)V,
\]

where \( Q \), \( K \), and \( V \) denote the query, key, and value matrices computed from the input embeddings, and \( d_k \) is the dimension of the key vectors, which influences the scaling of the dot products.

The transformer blocks encapsulate this self-attention mechanism, alongside multi-layer perceptrons (MLP), layer normalization (LN), and skip connections, to form the encoder:

\begin{align*}
&\text{LN}(\text{SelfAttention}(\text{LN}(x)) + x), \\
&\text{LN}(\text{MLP}(\text{LN}(x)) + x).
\end{align*}

For classification tasks, the ViT architecture typically prepends an additional token, the $\text{\texttt{[CLS]}}$ token, to the embedded patches. During the forward pass, the $\text{\texttt{[CLS]}}$ token interacts with patch embeddings via self-attention mechanisms in each transformer layer, thereby integrating global contextual information from all patches. The evolution of this token through the layers allows it to serve as a distilled representation of the image, encoding the aggregate information necessary for classification. In our experiments, we use the embedded $\text{\texttt{[CLS]}}$ token as the feature representation and subsequently as the input to the classification head.

\section{Benchmarks}
To ensure a thorough evaluation, we employ a diverse set of benchmarks, particularly focusing on those that present significant variations from the characteristics of ImageNet (i.e. the pretraining data of the backbone). In the following we provide a description of the key characteristics of each benchmark. 

\textbf{CIFAR-100}: The CIFAR-100 \citep{cifar100} dataset is a widely recognized benchmark in the field of machine learning for evaluating image classification performance. It consists of 60,000 color images, divided into 100 classes, with each class containing 600 images. Of these, 500 images per class are designated for training, and 100 images per class are set aside for testing. This dataset challenges pretrained models with its low-resolution images (32x32 pixels), which differ from the higher resolution images found in ImageNet. CIFAR-100 is frequently used as a continual learning benchmark \citep{wang2022l2p, wang2022dualprompt, gao2023lae, zhou2023revisiting} by splitting the classes into non-overlapping tasks. Our experiments use increments of 10 classes resulting in 10 tasks.

\textbf{ImageNet-R}: ImageNet-R (Rendition) \citep{imagenetr} extends the diversity of the original ImageNet dataset by including a variety of artistic renditions, such as paintings, sculptures, and sketches, across 200 classes. It comprises over 30,000 images, aiming to evaluate the robustness of models to visual variations not covered by standard object recognition datasets. This benchmark tests the ability of models to generalize across different styles and representations. For the class-incremental learning setting, we split the available classes into 10 tasks of 20 classes each, as done by \cite{wang2022l2p, wang2022dualprompt, gao2023lae}.

\textbf{ImageNet-A}: ImageNet-A \citep{imageneta} tests the robustness of models against natural adversarial examples. It consists of images often misclassified by pre-trained models, highlighting the challenges posed by real-world visual anomalies. The dataset includes around 7,500 images from different classes, focusing on instances that differ significantly from the training data distribution in ImageNet. This helps to test the adaptability of models to unexpected variations. In the continual learning setting, we group the available classes into 10 tasks of 20 classes each.

\textbf{DomainNet}: DomainNet \citep{domainnet} is a large-scale dataset aimed at evaluating domain adaptation techniques. It features approximately 600,000 images across 345 categories, spanning six distinct domains (clipart, infograph, painting, photo, quickdraw, and sketch). This dataset challenges pretrained models with significant domain shifts, requiring effective adaptation strategies to maintain high performance across different visual representations. We follow \cite{gao2023lae} and split the 345 classes into 5 increments of 69 classes each.

\textbf{ObjectNet}: ObjectNet \citep{objectnet} focuses on assessing object recognition models under varying conditions, such as changes in background, rotation, and viewpoint. It includes around 50,000 images, distributed across 313 object classes. Unlike ImageNet, the images in ObjectNet are collected with minimal bias, offering a realistic benchmark for testing the generalization capabilities of pretrained models to real-world scenarios. Following \cite{zhou2023revisiting}, we select a subset of 200 classes, which we split into 10 tasks of 200 classes each.

\textbf{Omnibench}: Omnibench \citep{omnibench} is a benchmark suite designed to evaluate models across a broad range of tasks, including image classification, object detection, and more. The benchmark contains 21 semantic realm-wise datasets that have no overlapping concepts. It aims to provide a comprehensive assessment of model performance and adaptability, addressing the limitations of benchmarks that focus on a single task or domain. From the original 7,372 classes, we sample 300 categories to construct the class-incremental learning set consisting of 10 tasks with 30 classes each (as done by \cite{zhou2023revisiting}). 

\textbf{CUB-200}: The CUB-200 \citep{cub200-2011} dataset, short for Caltech-UCSD Birds 200, is a fine-grained classification benchmark consisting of 11,788 images of 200 bird species. Each species includes a number of images for training and testing, challenging models with the task of distinguishing between highly similar categories based on subtle visual cues. We split the 200 available classes into 10 tasks of 10 classes for our continual learning setting.

\textbf{Cars196}: The Cars \citep{cars196} dataset contains 16,185 images of 196 classes of cars, ranging from sedans to SUVs. The dataset is split into a training set and a testing set, with the aim of evaluating models on fine-grained image classification tasks. The high level of similarity among classes poses a significant challenge, testing the precision of the pretrained models. For our experiments, we split the available classes into increments of 14 resulting in 14 tasks,

\textbf{Caltech256}: Caltech256 \citep{caltech256} is a dataset designed for object recognition tasks, containing 30,607 images spread across 256 object categories, plus a background/clutter category. Each category has at least 80 images, providing varied visual examples for training and testing models. To create uniformly sized increments we omit the clutter class and split the remaining classes into 8 tasks of 32 classes each.

\section{Implementation Details}
The implementation details of the approaches tested are founded on the codebase of \cite{gao2023lae}, which serves as the groundwork for both data loading, utilizing the \textit{continuum} \citep{douillardlesort2021continuum} library, and the training framework. To facilitate a uniform evaluation, the classification head across all approaches remains consistent, as defined within the LAE codebase \citep{gao2023lae}, establishing an equitable baseline for comparison.

\textbf{Query-based P-RFCL Approaches}: The Learning to Prompt (L2P) \citep{wang2022l2p} and DualPrompt (DP) \citep{wang2022dualprompt} methods are integrated into the codebase from their respective PyTorch implementations. Similarly, CODA \citep{smith2023coda} was incorporated directly from its official repository. To maintain fidelity to the original methods, the experiments adhere to the hyperparameter configurations recommended by the authors of these approaches. We choose the smaller version (CODA-P-S) for CODA to allow for a fair comparison to other methods (e.g. L2P or DP) as the standard CODA version surpasses the other approaches by a magnitude ($>3$M trainable parameters).

\textbf{PEFT-based Approaches}: We employ the LAE-adapter \citep{gao2023lae} methodology in its original form. Additionally, ADAM-based methods and the SimpleCIL baseline \citep{zhou2023revisiting} are integrated from their primary repositories, ensuring that each approach is represented accurately within the unified codebase.

\textbf{Simple P-RFCL Approach}: We have implement the simple P-RFCL baseline approach \textit{OnlyPrompt} in our codebase. To improve its performance, we employ a dual learning rate strategy that separates the learning rates of the classification head from the prompts. Our experiments show that tuning the learning rates of these components separately results in better performance.

The training epochs for all benchmarks are adjusted to ensure model convergence for each task. For larger or less complex benchmarks such as CIFAR100, DomainNet, ObjectNet, and Caltech256, each approach is trained for 5 epochs per task. However, for smaller or more complex benchmarks such as ImageNet-A, Omnibench, ImageNet-R, CUB200, and Cars196, we use 20 or 50 epochs per task. As suggested by \cite{gao2023lae}, we use a lower learning rate on benchmarks with 50 epochs per task. 

To optimize performance, we apply further refinements to the learning rate settings for the linear classification head and the OnlyPrompt method. We find that using a lower learning rate for the classification head $\text{lr}_{\text{head}} \in \{0.001, 0.0001\}$ and a higher learning rate for the prompt tokens $\text{lr}_{\text{prompt}} \in \{0.1, 0.01\}$ works best. Please note that the optimal learning rate in the OnlyPrompt setting is also dependent on the number of adaptation parameters, as the learning rate affects the adaptation-forgetting trade-off discussed in Section \ref{subsec:uniprompt}.

When implementing additional regularization techniques such as EWC or SI to the OnlyPrompt method, a thorough parameter search is performed. This search is customized for each benchmark and employs a validation set to optimize the regularization strength (lambda). This helps ensure that the regularization effect is stable across different datasets, thereby increasing its robustness.

\section{Additional Experimental Results}
In the HiDe Prompt approach \citep{wang2023hierarchical}, Task-Adaptive Prediction (TAP) is introduced to align classifier heads for individual tasks. This involves generating artificial pseudo embeddings from summary statistics of each task’s samples, which are then used to train and align the classifier heads of previous tasks. To evaluate TAP’s effectiveness, we conducted experiments comparing HiDe Prompt with and without TAP against our OnlyPrompt baseline. This comparison provides a fair assessment of TAP’s impact on both methods.

\begin{table*}[t]
    \centering
    \caption{Comparison of HiDe Prompt to our OnlyPrompt method with and without Task-Adaptive Prediction (TAP).}
    \resizebox{\textwidth}{!}{%
    \begin{tabular}{lcccccccccc|c}
        \toprule
        \textbf{Adaptation} & \textbf{Params (K)} & \textbf{CIFAR100} & \textbf{ImageNet-R} & \textbf{ImageNet-A} & \textbf{Domainnet} &  \textbf{ObjectNet} & \textbf{Omnibench} & \textbf{CUB} & \textbf{Cars} & \textbf{Caltech256} & \textbf{Avg.}\\
        \midrule
        HiDe (/wo TAP) & 384 & 85.3 ($\pm$0.8) & 75.8 ($\pm$0.8) & 44.7 ($\pm$1.7) & 66.0 ($\pm$0.2) & 59.9 ($\pm$0.8) & 67.9 ($\pm$1.0) & 82.7 ($\pm$0.6) & 47.4 ($\pm$11.4) & 94.3 ($\pm$0.1) & 69.3 ($\pm$11.7) \\
        HiDe (/w TAP) & 384 & 88.2 ($\pm$0.4) & \textbf{76.3} ($\pm$0.6) & 48.1 ($\pm$0.6) & \textbf{66.9} ($\pm$0.3) & 60.7 ($\pm$0.9) & \textbf{75.9} ($\pm$1.1) & \textbf{87.4} ($\pm$1.0) & \textbf{57.6} ($\pm$2.2) & \textbf{95.0} ($\pm$0.1) & \textbf{72.9} ($\pm$3.0) \\
        \midrule
        \textit{OnlyPrompt} (/wo TAP) & 3.84 & 87.0 ($\pm$0.5) & 72.1 ($\pm$1.8) & 55.0 ($\pm$0.6) & 66.4 ($\pm$0.3) & \textbf{61.4} ($\pm$0.1) & 70.1 ($\pm$1.1) & 81.4 ($\pm$1.8) & 46.6 ($\pm$7.0) & 94.4 ($\pm$0.1) & 70.5 ($\pm$7.6) \\
        \textit{OnlyPrompt} (/w TAP) & 3.84 & \textbf{88.3} ($\pm$0.3) & 74.3 ($\pm$0.6) & \textbf{57.5} ($\pm$1.0) & 63.1 ($\pm$0.1) & 60.0 ($\pm$0.6) & 75.4 ($\pm$0.5) & 84.8 ($\pm$1.0) & 37.2 ($\pm$2.0) & 94.7 ($\pm$0.2) & 70.6 ($\pm$2.7) \\
        \bottomrule
    \end{tabular}}
    \label{tab:tap_comp}
\end{table*}

The results, shown in Table \ref{tab:tap_comp}, indicate that HiDe Prompt without TAP is outperformed by our OnlyPrompt method on most benchmarks. However, when the TAP alignment strategy is applied, HiDe Prompt’s performance improves, surpassing OnlyPrompt without TAP on several benchmarks. Since the TAP alignment strategy is not specific to HiDe Prompt and only affects the training of classification heads, we incorporated it into our OnlyPrompt approach. Using this combined strategy, we enhanced OnlyPrompt’s performance, for example, from 70.1\% to 75.4\% on Omnibench, thereby matching the performance of HiDe Prompt with TAP on most benchmarks. The largest performance gains through TAP were observed on the Omnibench, CUB, and Cars benchmarks, aligning with the insights discussed in Section \ref{subsec:full_vs_single_adapt} regarding classifier misalignment for fine-grained classes.

\end{document}